\pdfoutput=1

\documentclass[11pt]{article}

\usepackage[]{acl}
\usepackage{times}
\usepackage{latexsym}

\usepackage[T1]{fontenc}

\usepackage[utf8]{inputenc}

\usepackage{hyperref}       
\usepackage{url}            
\usepackage{booktabs}       
\usepackage{amsfonts}       
\usepackage{nicefrac}       
\usepackage{microtype}      
\usepackage{xcolor}         
\usepackage{algorithm}      
\usepackage{blindtext}
\usepackage{graphicx}
\usepackage{amsmath}
\usepackage{multirow}
\usepackage{enumitem}
\title{Cognitive Visual-Language Mapper: Advancing Multimodal Comprehension with Enhanced Visual Knowledge Alignment}

\author{Yunxin Li$^{1}$, Xinyu Chen$^{1}$, Baotian Hu$^{1}$\thanks{~~~Corresponding author.}, Haoyuan Shi$^{1}$, Min Zhang$^{1}$\\
$^{1}$Harbin Institute of Technology, Shenzhen, China\\
\texttt{liyunxin987@163.com}
\\
\texttt{\{hubaotian, zhangmin2021\}}@hit.edu.cn
}

\begin{document}
\maketitle
\begin{abstract}


Evaluating and Rethinking the current landscape of Large Multimodal Models (LMMs), we observe that widely-used visual-language projection approaches (e.g., Q-former or MLP) focus on the alignment of image-text descriptions yet ignore the visual knowledge-dimension alignment, i.e., connecting visuals to their relevant knowledge. Visual knowledge plays a significant role in analyzing, inferring, and interpreting information from visuals, helping improve the accuracy of answers to knowledge-based visual questions. In this paper, we mainly explore improving LMMs with visual-language knowledge alignment, especially aimed at challenging knowledge-based visual question answering (VQA). To this end, we present a \textit{Cognitive Visual-Language Mapper} (CVLM), which contains a pretrained Visual Knowledge Aligner (VKA) and a Fine-grained Knowledge Adapter (FKA) used in the multimodal instruction tuning stage. Specifically, we design the VKA based on the interaction between a small language model and a visual encoder, training it on collected image-knowledge pairs to achieve visual knowledge acquisition and projection. FKA is employed to distill the fine-grained visual knowledge of an image and inject it into Large Language Models (LLMs). We conduct extensive experiments on knowledge-based VQA benchmarks and experimental results show that CVLM significantly improves the performance of LMMs on knowledge-based VQA (average gain by 5.0\%). Ablation studies also verify the effectiveness of VKA and FKA, respectively.\footnote{Codes and Data are available at \url{https://github.com/HITsz-TMG/Cognitive-Visual-Language-Mapper}}

\end{abstract}

\section{Introduction}

\begin{figure}[t]
    \centering
    \includegraphics[width=0.49\textwidth]{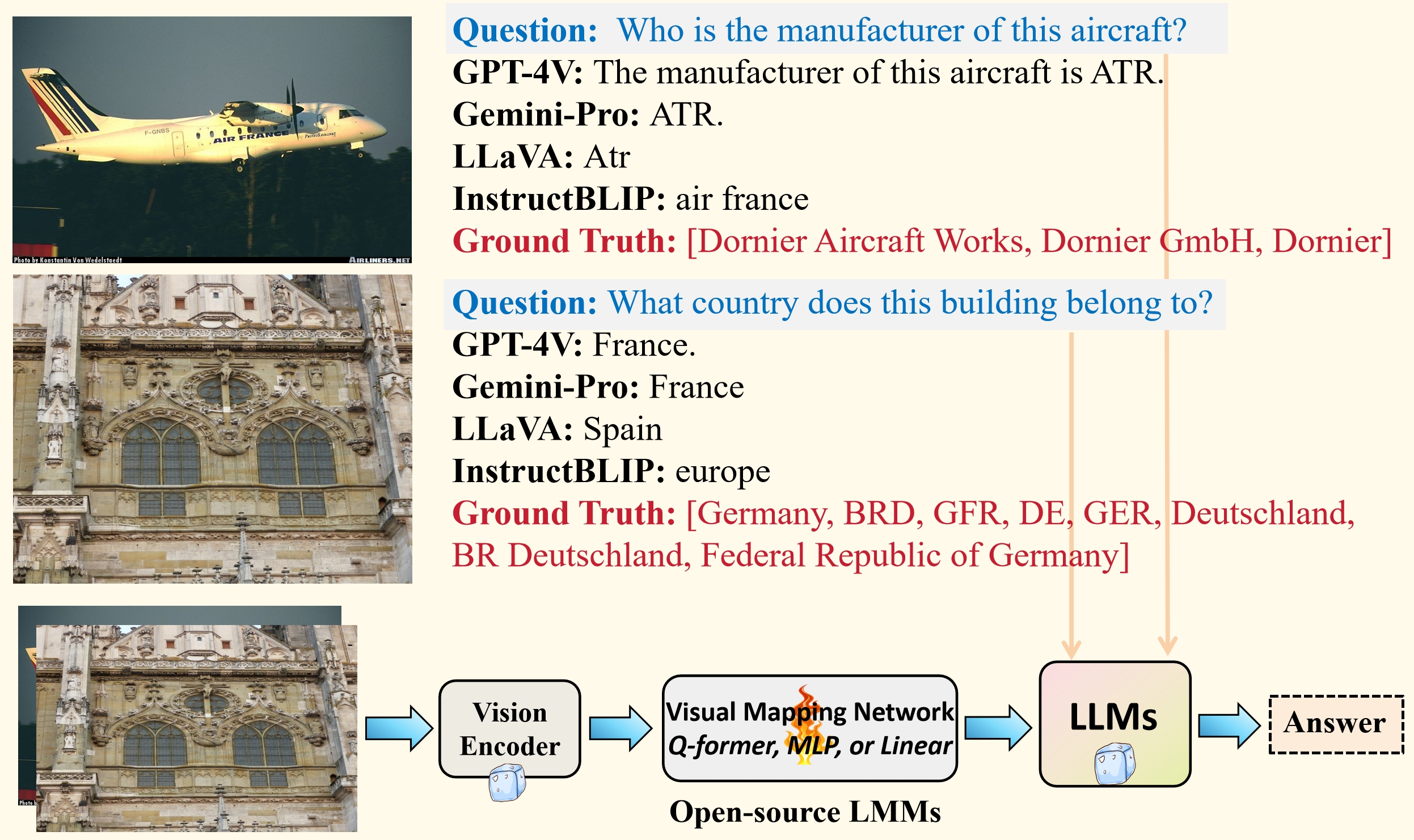}
    \caption{It illustrates the performance of LMMs on visual information-seeking questions. The bottom part shows the widely-used architecture of open-source LMMs, where the visual mapping network is usually pretrained on massive image-text captioning data. All LMMs including GPT-4V (Date: 2023.11.17) and Gemini-Pro make incorrect decisions. }
    \label{fig:intro_case}
\end{figure}

Recent Large Multimodal Models (LMMs) such as GPT-4V~\cite{gpt4}, Gemini~\cite{team2023gemini}, MiniGPT-4~\cite{zhu2023minigpt}, InstructBLIP~\cite{liu2023visual}, LLaVA~\cite{liu2023improved}, and many others, have achieved impressive performance in a variety of visual understanding and reasoning tasks, especially on Visual Question Answering (VQA)~\cite{li2023comprehensive, li2023lmeye}.
Current open-source LMMs are usually constructed by combining pertained visual encoders and Large Language Models (LLMs), as depicted in the bottom part of Figure~\ref{fig:intro_case}, where a visual mapping network (e.g., Q-former~\cite{li2023blip2}, Linear~\cite{zhu2023minigpt}, or MLP~\cite{liu2023improved,li2023multi}) is employed to project visual representations into the language space of LLMs. Although such LMMs have achieved powerful visual understanding capability similar to GPT-4V and Genimi on some image understanding tasks such as Image Captioning~\cite{changpinyo2021conceptual}, Visual Dialogue~\cite{zhang2022reasoning, chen2022utc}, Visual Entailment~\cite{xie2019visual, do2020snli}, and VQA~\cite{VQA}, they often fall short of knowledge-based VQA, which necessitates relevant knowledge to answer these visual questions. As the cases illustrated in Figure~\ref{fig:intro_case}, these advanced LMMs (including GPT-4V and Gemini-Pro) can not give correct answers to simple visual information seeking questions: \textit{Who is the manufacturer of this aircraft; What country does this building belong to?}.

In light of this, rethinking the construction process of LMMs~\cite{selfinstruct, li2023lmeye, zhu2023minigpt, liu2023improved,koh2023grounding} from the initial pretraining stages, we discover that these visual mapping networks trained on massive image-text captioning pairs simply transfer visual features to their language descriptions. They overlook the visual language knowledge-dimension alignment. i.e., connecting visuals to their relevant knowledge. As we know, visual knowledge~\cite{Collins2014} plays a pivotal role in the way humans understand and interact with the world. It extends beyond the mere ability to recognize and interpret visuals, incorporating an understanding of spatial relationships, patterns, and symbols, which are essential components of human cognition~\cite{pinker1984visual, cavanagh2011visual}. Additionally, previous works also demonstrated that introducing visual knowledge~\cite{lu2022imagination,zhu2022visualize, li2023multi,li2023towards} can improve the performance of pretrained language models on natural language understanding~\citep{lu2022imagination} and open-ended text generation tasks~\citep{zhu2022visualize}. 
Inspired by these insights, we focus on enhancing LMMs through the introduction of visual-language knowledge alignment, going beyond the conventional scope of visual-language integration.

To this end, we present a \textit{Cognitive Visual-Language Mapper} (\textbf{CVLM}), which contains a pretrained Visual Knowledge Aligner (VKA) and a Fine-grained Knowledge Adapter (FKA). 
Specifically, we devise VKA based on a small language model that interacts with fine-grained image representation in each block. The output hidden states of VKA are fed into the LLM as the knowledge embedding tokens by a linear projection layer. To make VKA effectively capture image-relevant knowledge, we first train it on image-knowledge pairs~\cite{srinivasan2021wit} collected from Wikipedia~\footnote{https://en.wikipedia.org/wiki/Wikipedia:Images} via the next tokens prediction. Like Q-former~\cite{li2023blip2} and prefix-tuning~\cite{li2021prefix}, we only fine-tune some learnable query tokens and the linear layer to acquire fixed-length visual knowledge representation and convert it into the representation space of LLM. In addition,
considering that visual objects contain fine-grained visual knowledge, we introduce FKA to gain comprehensive visual knowledge of an image and distill valuable visual knowledge from the whole knowledge representation sequence. The output knowledge vectors of FKA are injected into each layer of LLMs to realize in-depth interactions between LLMs and detailed visual knowledge. By doing so, CVLM is capable of connecting visuals to relevant knowledge, enabling LMMs to utilize them during multimodal understanding and generation. 


To verify the effectiveness of CVLM, we conduct extensive experiments on image-centered, knowledge-based, and complex visual reasoning question-answering scenarios: VQAv2~\cite{balanced_vqa_v2}, OKVQA~\cite{marino2019ok}, A-OKVQA~\cite{AOKVQA}, Infoseek~\cite{chen2023can}, TextVQA~\cite{singh2019towards}, and SeedBench~\cite{seedBENCH}. The experimental results show that CVLM significantly outperforms previous strong baselines such as LLaVA-v1.5. The ablation and case studies indicate that CVLM is capable of linking visual knowledge and improving performance on knowledge-intensive tasks via the introduced aligner and adapter.

Our contributions can be summarized as follows:
\begin{itemize}
    \item We present a cognitive visual-language mapper to achieve visual-language knowledge alignment, which contains a pretrained visual knowledge aligner and a fine-grained knowledge adapter that is used to distil and inject valuable visual knowledge into LLMs. 

    \item To the best of our knowledge, we are the first to explore the visual-language knowledge alignment during the pretraining and finetuning stages of LMMs, connecting visuals to their knowledge via CVLM.

    \item Experimental results indicate that CVLM significantly improves the performance of LMMs on knowledge-intensive VQA. The ablation studies also verify the effectiveness of VKA and FKA on specific knowledge-based VQA.
\end{itemize}

\begin{figure*}[t]
    \centering
    \includegraphics[width=1.0\textwidth]{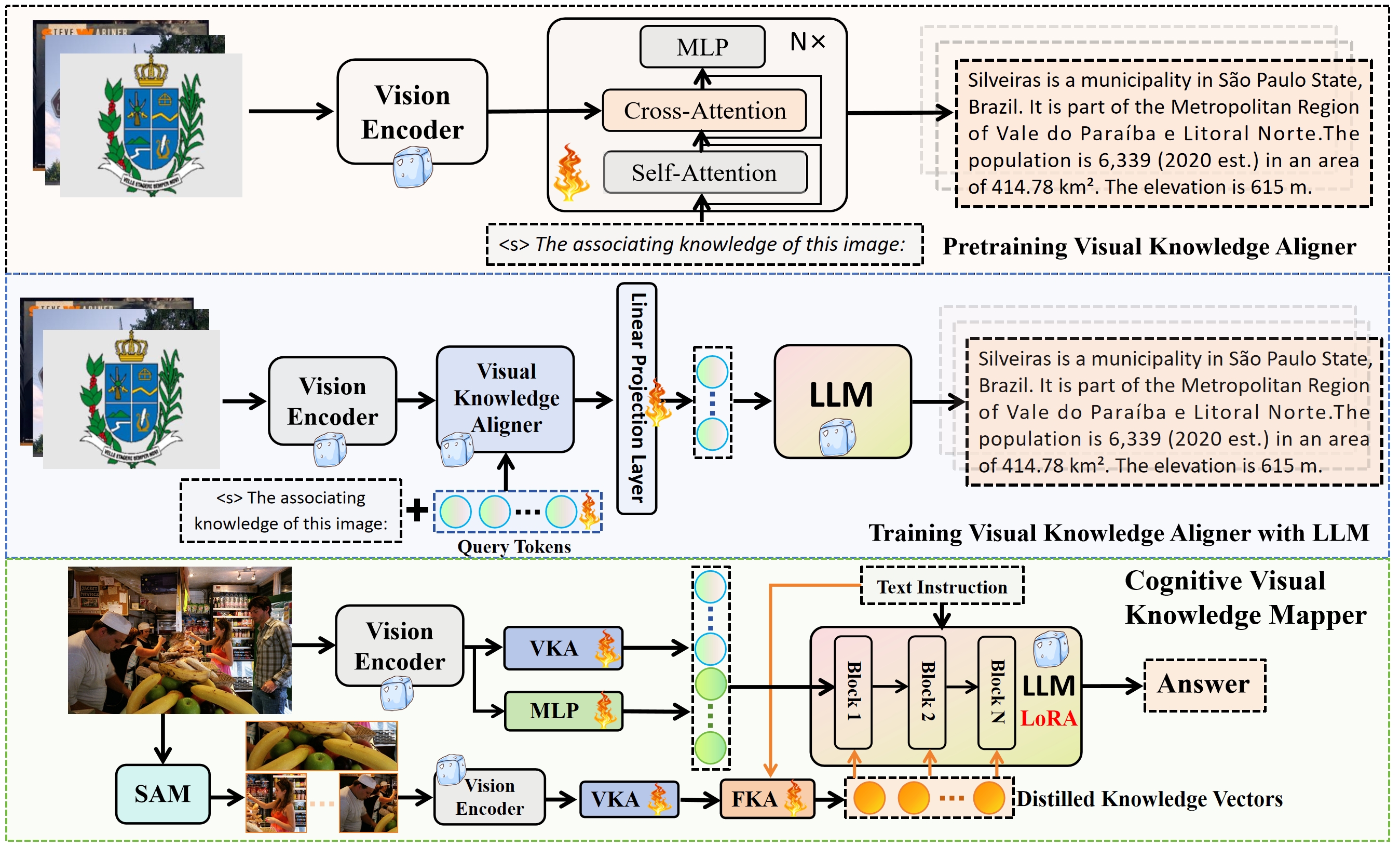}
    \caption{An overview of Cognitive Visual Knowledge Mapper. From top to bottom, it shows 1) Pretraining visual knowledge aligner, where we use a pretrained small language model to interact with image features via the cross attention module; 2) Training visual knowledge aligner with LLM, in which we realize visual knowledge alignment between vision encoder and LLM via the learnable query tokens and linear layer; 3) Overall architecture of CVLM, where we present the fine-grained visual knowledge adapter beyond common visual projection (MLP) and VKA. }
    \label{fig:model}
\end{figure*}

\section{Related Work}

\textbf{Knowledge-based Visual Question Answering}. 
Visual Question Answering (VQA) is a multidisciplinary field that combines vision and language processing to address queries about images. A recent development in this domain is knowledge-based VQA, which relies on external information for open-domain visual questions. The initial knowledge-based VQA datasets, KB-VQA~\cite{wang2015explicit} and FVQA~\cite{wang2017fvqa} had limited knowledge requirements, referred to as "closed" knowledge.
In contrast, S3VQA~\cite{jainSelect} and OK-VQA~\cite{marino2019ok} datasets introduced questions demanding "open-domain" knowledge, incorporating widely recognized facts from diverse domains. INFOSEEK~\cite{chen2023can}, a recent Wikipedia-based VQA dataset, concentrated on fine-grained entity knowledge for open-domain information-seeking queries.
As a result, datasets like OK-VQA and INFOSEEK, encompassing diverse knowledge categories, are ideal for assessing the performance of LMMs in open-domain VQA tasks. A-OKVQA~\cite{AOKVQA} necessitates a broad foundation of common sense and worldly knowledge for answering questions, like QA on knowledge graph~\cite{chen-etal-2023-multi,chen2023temporal}.

\textbf{Large Visual-Language Models}.
Recent advancements in foundational models for vision and language have led to the development of LMMs. In response to large model GPT-4~\cite{gpt4}, several others have emerged, including GVT~\cite{GVT2023}, MPlug~\cite{ye2023mplug}, Macaw~\cite{lyu2023macaw}, LMEye~\cite{li2023lmeye}, and LLaVA~\cite{liu2023improved}. These models have demonstrated strong performance across various visual-language tasks.
Typically, these models utilize pretrained visual models to extract visual features, which are then integrated into the linguistic space of LMMs through a straightforward projection layer. This layer can be a Linear Layer~\cite{merullo2022linearly, liu2023visual, li2023multi} or a Q-former~\cite{li2023blip2}. Following this integration, similar to the supervised fine-tuning approach used for LLMs, these systems undergo refinement using diverse and high-quality multimodal instruction-following datasets~\cite{liu2023visual, zhu2023minigpt, ye2023mplug}. These datasets encompass both human-labeled data for downstream tasks like Visual Question Answering (VQA) and VCR~\cite{VQA, zellers2019recognition}, as well as datasets automatically generated by GPT-4. Meanwhile, some multimodal benchmarks like MMBench~\cite{liu2023mmbench} and SEED-Bench~\cite{seedBENCH} have been established to evaluate advanced LMMs, and \citet{li2023comprehensive_gpt4v} presents a comprehensive assessment of their performance in knowledge-intensive VQA scenarios. 

\section{Methodology}

\subsection{Overview}

CVLM focuses on connecting visuals to relevant knowledge and injecting them into LLMs to perform multimodal instruction-following generation. The overview of CVLM is illustrated in Figure~\ref{fig:model}. Specifically, given an image $I$ and text instruction $T=(t_1,t_2, ..., t_M)$, where $t_M$ refers to the $M$ th token of instruction,  
we initially utilize a visual encoder to generate representations of images. These representations are then mapped into the language space of LMMs using a Multilayer Perceptron (MLP). We integrate a Visual Knowledge Aligner (VKA) between the frozen visual encoder and LLMs to transform the visual knowledge into the language space of LLMs. 
Furthermore, acknowledging that image regions detailed visual knowledge, we introduce a Fine-Grained Visual Knowledge Adapter (FKA). This adapter is designed to extract valuable information from the intricate visual knowledge representations produced by the VKA. This distilled knowledge is subsequently incorporated into the LLMs. Through this methodology, we enable the LLMs to not only associate with but also utilize visual knowledge effectively, thereby facilitating multimodal generation in an end-to-end fashion.

\subsection{Visual Knowledge Aligner}
Firstly, we introduce a task-agnostic visual knowledge generator to realize associating relevant visual knowledge given an image. Specifically, we employ a pretrained visual encoder CLIP ViT-L/14 with inputting image size of 336*336 to gain the image representation sequence $\mathbf{h}_I = (h_g, h_1^{I}, ...., h_{576}^{I})$, where $h_g$ and $h_{576}^{I}$ refers to the global feature of the image and $576$ th patch representations. Then, as shown in the top part of Figure~\ref{fig:model}, we utilize a pretrained small autoregressive language model (OPT-1.3B) as the generator of visual knowledge, which interacts with the visual sequence $h_I$ via adding the cross attention layer in each block. We train it on an amount of image-knowledge pairs via the next token prediction. 
These pairs are meticulously curated from Wikidata, ensuring a rich and diverse source of world knowledge information.
This pretrained knowledge generator is capable of associating relevant knowledge based on an input image. 

Afterward, as depicted in the middle part of Figure~\ref{fig:model}, we use the pretrained visual knowledge generator as the backbone to construct the whole visual-knowledge aligner like Q-former. Specifically, we add learnable tokens $\mathbf{h}_{\text{KQ}} = (h_{\text{KQ}}^1, ..., h_{\text{KQ}}^{N})$, where $N$ refers to the number of query tokens, which is joined with the knowledge prompt ``\textit{<s> The associating knowledge of this image}''. We adopt a learnable linear projection layer to project the obtained features into the language space. The whole process could be presented in Eq. \ref{eq1}.
\begin{equation}
    \begin{array}{c}
        \mathbf{h}_{\text{AO}} = \text{VKA} ([\mathbf{h}_{\text{KP}}, \mathbf{h}_{\text{KQ}}], \mathbf{h}_I), \vspace{1.0ex}\\
        \mathbf{h}_{\text{KO}} = \mathbf{W}^{K}\mathbf{h}_{\text{AO}} + \mathbf{b},
    \end{array}
\label{eq1}
\end{equation}
where $\mathbf{h}_{\text{AO}}$ and $\mathbf{h}_{\text{KP}}$ show the output of the pretrained visual knowledge generator and the word embeddings of knowledge prompt, respectively. $[,]$ refers to the sequence concatenation of two vectors. $\mathbf{W}^{K}\in \mathbf{R}^{d_{K} \times d_{L}}$ and $\mathbf{b} \in \mathbf{R}^{d_{L}}$ are the learnable parameters, where $d_{K}$ and $d_{L}$ represent the hidden state dimensions of visual knowledge aligner and LLM, respectively. $\mathbf{h}_{KO}$ will be fed into the language models with the original image representation $\mathbf{h}_{\text{IO}}$. It is gained by a learnable MLP trained on image-text captioning pairs, i.e., $\mathbf{h}_{\text{IO}} = \text{MLP}([h_1^{I}, ...., h_{576}^{I})])$.

The supervision signal remains the sequence of knowledge relevant to the image and the learning objective is the cross-entropy generation loss.  By doing so, the designed knowledge aligner could connect visuals to their knowledge and project them into the LLMs.

\subsection{Fine-grained Visual Knowledge Adapter}

\begin{figure}[t]
    \centering
\includegraphics[width=0.5\textwidth]{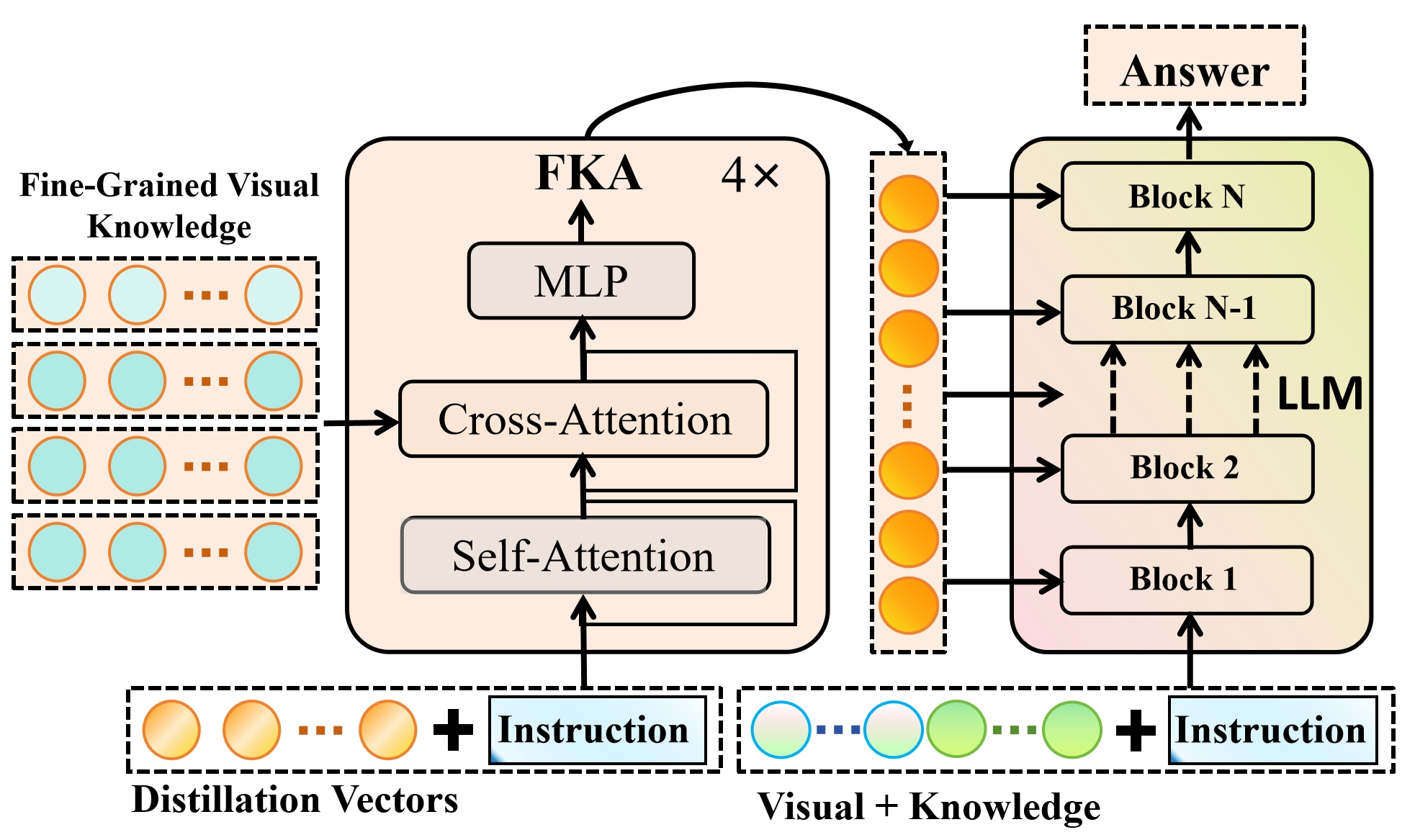}
    \caption{The detailed calculation process of fine-grained visual knowledge adapter, i.e., VKA shown in Figure~\ref{fig:model}. ``Visual + Knowledge'' indicates the representation concatenation of an image $\mathbf{h}_{\text{IO}}$ and its relevant knowledge projection $\mathbf{h}_{\text{KO}}$. }
    \label{fig:fka_intro}
\end{figure}

Considering that image regions (e.g., objects) also link to useful knowledge, we present a fine-grained visual knowledge adapter (FKA) to gain and distill detailed and useful visual knowledge, which is injected into each block in LLMs.
Firstly, we obtain the fine-grained visual knowledge representations by using widely used segment anything tools named SAM~\cite{kirillov2023segment} and VKA. Concretely, as shown in the bottom-left part of Figure~\ref{fig:model}, we use the SAM to obtain the image regions with their confidence scores and adopt the top five image objects, which could be denoted as ${I_1, ..., I_5}$. Then, we utilize vision encoder and VKA to obtain the fine-grained visual knowledge representations $\mathbf{h}_{I_1},\mathbf{h}_{I_2},...,\mathbf{h}_{I_5}$, which are illustrated in Figure~\ref{fig:fka_intro}. 

Subsequently, we employ a four-layer transformer decoder with learnable distillation vectors $\mathbf{h}_D = (\mathbf{h}_{1}^{D}, ..., \mathbf{h}_{N}^{D})$, where N is the number of distill vectors. To gain useful knowledge, we splice text instruction $T$ after $\mathbf{h}_D$ to distill instruction-relevant visual knowledge. 
The specific calculation progress of each FKA block is given in Eq.~\ref{eq2}
\begin{equation}
    \begin{array}{c}
         \mathbf{h}_{S}^{l-1} = \text{Self-A} (\mathbf{h}_D, \mathbf{h}_T) + \mathbf{h}_{\text{FKA}}^{l-1}, \vspace{1.0ex}\\
        \mathbf{h}_{cr}=\text{Cross-A}(\mathbf{h}_{S}^{l-1}, [\mathbf{h}_{I_1},\mathbf{h}_{I_2},...,\mathbf{h}_{I_5}]) + \mathbf{h}_{S}^{l-1}, \vspace{1.0ex}\\
        \mathbf{h}_{\text{FKA}}^{l} = \text{MLP}(\mathbf{h}_{cr}),
    \end{array}
\label{eq2}
\end{equation}
where  $\mathbf{h}_{\text{FKA}}^{l-1}$ and $\mathbf{h}_T$ represent the output of the previous block of FKA and textual instructions, respectively. \text{Self-A} and \text{Cross-A} refers to the self and cross attention calculation in Transformer. Through the whole calculation of the FKA module, we can obtain the distilled visual knowledge $\mathbf{h}_{\text{FKA}}$ and will inject it into each block in LLMs to achieve in-depth interaction between LLMs and fine-grained visual knowledge. 

As the right part shown in Figure~\ref{fig:fka_intro}, the input of the first layer in LLMs is $\mathbf{h}_{\text{IO}}, \mathbf{h}_{\text{KO}}, \text{and}~\mathbf{h}_{T}$. We splice the sequence $\mathbf{h}_{\text{FKA}}$ according to the depth of LLMs and obtain the sequence of injected vectors: $\mathbf{h}_{\text{FKA}}^{1},..., \mathbf{h}_{\text{FKA}}^{\textit{LN}}$, where \textit{LN} represents the total number of layers for LLMs. The length of injected vectors is equal to $\textit{N}/\textit{LN}$ and it will be spliced to the front of the input sequence. In summary, the whole framework is capable of visual knowledge alignment via VKA at the pretraining stage and the efficient injection of fine-grained visual knowledge via FKA.

\subsection{Training}

We pre-train the backbone of VKA and train VKA with the LLM on the same image-knowledge pairs from Wikipedia-based image text dataset WIT~\cite{srinivasan2021wit} via the cross-entropy loss. Suppose that the training target is $Y = (y_1,...,y_{\text{KM}})$, in which KM is the token length of knowledge description, the training objects are presented as the following two equations:
\begin{equation}
    \begin{array}{c}
    \mathcal{L}_{\text{VKA}}^{P} =  -\sum_{i=1}^{\text{KM}}\mathbf{log}{P}_{i}(\hat{y}_{i} = y_i | h_I, h_{\text{KP}};\\ y_1,...,y_{i-1}),     \vspace{1.0ex}\\
    \mathcal{L}_{\text{VKA}}^{A} =  -\sum_{i=1}^{\text{KM}}\mathbf{log}{P}_{i}(\hat{y}_{i} = y_i | h_I, h_{\text{KP}}, h_{\text{KQ}};\\ y_1,...,y_{i-1}),
    \end{array}
\end{equation}
where $\mathcal{L}_{\text{VKA}}^{P}$ and $\mathcal{L}_{\text{VKA}}^{A}$ represents the loss of pretraining and aligning stages, respectively.
While training CVLM on the multimodal instruction dataset, the overall training object is shown in Eq.~\ref{eq4}
\begin{equation}
     \mathcal{L}_{\text{CVLM}} =  -\sum_{i=1}^{N_A}\mathbf{log}{P}_{i}(\hat{y}_{i} = y_i |I, T;\\ t_1,...,t_{i-1}),
\label{eq4}
\end{equation}
where $N_A$ and $t_i$ refer to the total token count and the $i$ th token of an answer.

\section{Experiments}

\begin{table*}[t]
\renewcommand\arraystretch{1.10}
\begin{center}
    \scalebox{0.66}{
    \begin{tabular}{c c |c| c c c c c c c}
        \toprule
        \textbf{Method} & LLMs & Avg. & OK-VQA & VQAv2 & A-OKVQA\textsuperscript{M} & A-OKVQA\textsuperscript{O} & TextVQA & InfoSeek & SEED-Bench\textsuperscript{S} \\
        \hline
        Promptcap~\cite{hu2022promptcap} & GPT-3 & - & -  &73.2 & 56.3 & - & - & - & - \\
        Prophet~\cite{yu2023prophet} & mPLUG & - &- & 76.6 & 64.7 & - & - & - & - \\
        VPD (55B)~\cite{hu2023visual} & PaLI & - &- & 84.7 & 62.7 & - & - & - & - \\ 
        Flamingo-9B & - & - &44.7 & 51.8 & - & - & - & - & - \\
        BLIP2 & Flan-T5-XXL & - & 45.9 & 65.2 & - & 53.71 & - & 10.67 & - \\
        MiniGPT4 & Vicuna-7B & - & 32.16 & 44.31 & - & - & - & 10.03 & 47.4 \\ 
        InstructBLIP & Flan-T5-XL & 50.68 & 48.30 & 70.14 & 76.68 & 61.05 & 30.46 & 10.30 & 57.80 \\
        InstructBLIP & Flan-T5-XXL & 50.83 & 46.91 & 70.11 & 78.34 & 62.01 & 29.79 & 8.36 & 60.29 \\
        InstructBLIP & Vicuna-7B & 50.00 & 57.36  & 74.77 & 45.07 & 67.86 & 33.09 & 10.05 & 58.8 \\
        Qwen-VL~\cite{Qwen-VL} & Qwen & 55.92 & 57.13	& 77.89	&70.04	&67.25 & 41.94&	15.21	& 62.41 \\
        LLaVA-v1.5$^\ddagger$ &  Vicuna-7B & 53.00 & 50.8 & 72.5 & 73.45 & 65.27 & 45.81 & 8.18 & 55.05 \\        
        \hline
        CVLM w/o ( FKA \& VKA) &  Vicuna-7B & 52.93 & 50.4 & 72.7 & 73.97 & 64.37 & 45.76 & 8.18 & 55.11 \\        
        CVLM w/o FKA & Vicuna-7B & 55.03 & 52.8 & 73.7 & 77.12 & 65.59 & 47.46 & 9.33 & 59.63 \\
        CVLM &  Vicuna-7B & 56.17 & 54.3 & 75.6 & 77.64 & 66.99 & 49.30 & 10.72 & 58.77 \\
        CVLM (3M IKPairs) w/o FKA & Vicuna-7B & 57.19 & 55.7 & 75.7 & 77.90 & 70.22 & 49.89 & 11.21 & 59.73\\
        CVLM (3M IKPairs, Objects=1) & Vicuna-7B & 57.92 & 56.90 &	76.32&78.69 & 70.48	& 50.21	& 10.01 &	59.65 \\
        CVLM (3M IKPairs, Objects=3) & Vicuna-7B & 58.19 & 57.17 &	76.40 & 79.21&	70.91&	50.32 &	10.42	&62.93\\
        CVLM (3M IKPairs Objects=5) & Vicuna-7B & 57.83 & 56.92 & 76.48 & 78.95 & 70.39 & 50.48 & 11.27 & 60.30 \\ 
        CVLM (3M IKPairs, Objects=8) & Vicuna-7B & 57.28 & 56.71 &	76.0&78.95 & 69.08	&	49.59&	10.78&	59.83\\
        CVLM &  Qwen-VL & \textbf{60.23} & 58.91	& 80.88	&82.71	&72.14 & 45.02& 	15.45	& 66.28 \\
        \bottomrule
    \end{tabular} 
    }
    \caption{\textbf{Comparison between different LMMs on knowledge-based VQA benchmarks.}
    With 7B parameters, CVLM achieves the best performance with the same training data. 
    ``$\ddagger$'' shows that we fairly use the same instruction tuning data to train the model. ``IKPairs'' represents the image-knowledge pairs used to train VKA and the initial version is trained with 2M pairs. ``Objects'' refers to the number of object regions used in FKA, which are obtained by SAM.
    Benchmark names are abbreviated due to space limits.
    A-OKVQA\textsuperscript{M}: Multi-Choice A-OKVQA~\cite{AOKVQA};
    A-OKVQA\textsuperscript{O}: Open-ended A-OKVQA~\cite{AOKVQA};
   TextVQA~\cite{singh2019towards};
    Infoseek~\cite{chen2023infoseek};
    SEED-Bench\textsuperscript{S}: SEED-Bench (Spatial)~\cite{li2023seed};
    }
    \label{tab:benchmark_result} 
\end{center}
\end{table*}

\subsection{Data sets}
Knowledge-based VQA is a task that requires reasoning with joint visual information, textual instructions, and outside knowledge. 
We mainly evaluate LMMs on the following relevant datasets.
\textbf{OK-VQA~\cite{marino2019ok}} is a visual question-answering dataset that requires methods that can draw upon outside knowledge to answer questions. It contains 9,009 training samples and 5,046 validation samples.
\textbf{A-OKVQA~\cite{AOKVQA}} is a knowledge-based visual multiple-choice question-answering benchmark that contains 17,056 training samples, 1,145 validation samples, and 6,702 testing samples.
\textbf{VQAv2~\cite{balanced_vqa_v2}} is a visual open-ended question-answering dataset where answering questions requires an integrated understanding of vision, language, and commonsense knowledge. It contains 443,757 training samples, 214,354 validation samples, and 447,793 test samples.
\textbf{TextVQA~\cite{singh2019towards}} is a task concerning reading and reasoning about text within images to answer questions related to them. The dataset comprises 34,602 training samples, 5,000 validation samples, and 5,734 validation samples.
\textit{We integrated the training sets of the datasets above to construct a total of about 100K multi-turn instruction data}. Additionally, we also introduce the comprehensive evaluation benchmark SEED-Bench~\cite{seedBENCH} and InfoSeek~\cite{chen2023infoseek} dataset to assess LMMs on comprehensive spatial understanding and fine-grained visual knowledge inferring.

\begin{table*}[t]
\renewcommand\arraystretch{1.15}
\tabcolsep=0.08cm
\centering
\small

\begin{tabular}{l|c|cccccccccccc}
\toprule
Model & Avg. & Building & Animal & Plant & Location & Food & OC & Facility & Vehicle & Objects & Sport & Other \\
\hline
MiniGPT-4 (Vicuna-7b) & 10.03 & 7.33 & 6.66 & 5.33 & 10.0 & 24.67 & 4.0 & 7.33 & 18.67 & 6.67 & 14.0 & 8.67\\
BLIP-2 (FlanT5-xxl) & 10.67 & 8.7 & 2.67 & 4.0 & 16.0 & 14.0 & 9.33 & 16.0 & 28.0 & 2.0 & 9.33 & 7.33 \\
InstructBLIP$^{\clubsuit}$ (Vicuna-13b) & 8.50 & 3.3 & 2.0 & 1.33 & 10.0 & 10.67 & 6.0 & 4.67 & 26.67 & 2.67 & 20.67 & 5.33\\
InstructBLIP$^{\clubsuit}$ (FlanT5-xxl) & 8.37 & 4.0 & 5.33 & 2.0 & 8.67 & 8.0 & 8.0 & 8.0 & 28.0 & 5.34 & 8.67 & 6.0 \\
LLaVA-v1.5-13b$^{\clubsuit}$ & 10.22 & 11.33 & 16.67 & 0.0 & 24.67 & 6.0 & 0.7 & 10.67 & 26.0 & 5.3 & 0.13 & 10.0\\

LLaVA-v1.5-7b$^{\ddagger\clubsuit}$ & 8.18 & 5.33 & 6.67 & 3.33 & 10.00 & 11.33 & 6.67 & 3.33 & 28.67 & 2.67 & 5.33 & 6.67\\
\hline
CVLM (LD=0) & 9.33 & 3.33 & 14.67 & 5.33 & 6.0 & 14.0 & 6.0 & 2.67 & 36.67 & 4.0 & 0.67 & 9.33\\
CVLM (LD=2) & 10.72 & 5.33 & 10.0 & 2.67 & 10.67 & 14.0 & 6.0 & 2.0 & 36.0 & 1.34 & 21.33 & 8.67\\
CVLM (LD=4) & 9.94 & 4.0 & 8.0 & 2.0 & 9.33 & 15.33 & 4.67 & 2.67 & 38.0 & 1.33 & 16.67 & 7.33\\
CVLM (LD=8) & 10.55 & 4.0 & 8.67 & 2.67 & 9.33 & 14.67 & 4.67 & 2.0 & 36.0 & 2.67 & 24.0 & 7.33\\
CVLM (3M IKPairs, LD=0) & 11.21 & 4.67 & 10.0 & 5.33 & 8.67 & 15.33 & 5.44 & 3.33 & 38.0 & 3.33 & 22.67 & 6.67 \\
CVLM (3M IKPairs, LD=2) & 11.27 & 4.67 & 10.67 & 4.67 & 8.67 & 15.33 & 5.33 & 3.33 & 38.0 & 3.33 & 22.67 & 7.33\\
CVLM-624K (LD=0) & 12.12 & 4.0 & 11.33 & 2.0 & 10.0 & 16.67 & 6.0 & 3.33 & 37.33 & 6.0 & 28.0 & 8.67\\
CVLM-624K (LD=2) & \textbf{12.30} & 4.67 & 11.33 & 2.67 & 9.33 & 16.67 & 6.0 & 4.0 & 38.67 & 6.67 & 27.33 & 8.0\\
\bottomrule
\end{tabular}
\caption{\textbf{Held-out testing results on InfoSeek with fine-grained world knowledge}. Baseline results and knowledge categories are reported by \citet{li2023comprehensive_gpt4v}.
``LD'' represents the length of distillation vectors used in FKA. ``LD=0'' is identical to ``w/o FKA''. 
 `OC'' refers to Organization and Company. ${\clubsuit}$ indicates that the corresponding LMM baseline is trained using the training sets of knowledge-intensive datasets: OK-VQA and A-OKVQA.}
\label{tab:infoseek_results}
\end{table*}
\begin{table*}[t]
    \centering
    \small
    \begin{tabular}{c|ccccc}
        \hline
        Model & MMbench & ScienceQA-I & PoPE-Adversarial & PoPE-Random & PoPE-Popular \\
        \hline
        CVLM w/o (FKA \& VKA) & 50.85 & 62.76 & 82.94 & 86.98 & 86.14 \\
        CVLM w/o FKA & 56.46 & 69.21 & 83.13 & 87.81 & 85.85 \\
        CVLM & \textbf{59.78} & 68.92 & 82.78 & 88.79 & 85.64 \\
        CVLM (3M IKPairs) w/o FKA & 57.06 & 69.96 & 82.32 & 85.75 & 84.26 \\
        CVLM (3M IKPairs) & \textbf{59.10} & 69.85 & 82.47 & 86.90 & 84.68 \\
        \hline
    \end{tabular}
    \caption{Comparison of different models across various common benchmarks. ScienceQA~\cite{lu2022learn} and PoPE~\cite{li2023evaluating} are used to evaluate the complex reasoning and hallucination recognition abilities of LMMs.}
    \label{tab:pope}
\end{table*}

\begin{table*}[t]
\renewcommand\arraystretch{1.20}
\tabcolsep=0.14cm
\centering
\small

\begin{tabular}{l|c|cccccccccccc}
\toprule
Model & Avg. & VT & BCP & OMC & SR & CF & GHLC & PEL & PA & ST & WC & Other \\
\hline
MiniGPT-4 (Vicuna-7b) & 29.31 & 28.67 & 31.03 & 26.0 & 28.0 & 25.33 & 38.21 & 22.67 & 29.33 & 29.23 & 31.25 & 34.0\\
BLIP-2 (FlanT5-xxl) & 39.06 & 30.67 & 34.48 & 38.0 & 40.67 & 34.0 & 42.28 & 39.33 & 41.33 & 44.62 & 50.0 & 40.67\\
InstructBLIP$^{\clubsuit}$ (Vicuna-13b) & 41.02 & 34.00 & 52.41 & 37.33 & 51.33 & 33.33 & 46.34 & 31.33 & 38.67 & 32.30 & 49.11 & 43.33\\
InstructBLIP$^{\clubsuit}$ (FlanT5-xxl) & 47.96 & 44.66 & 51.03 & 48.67 & 48.0 & 43.33 & 51.22 & 47.33 & 42.0 & 55.38 & 58.04 & 45.33 \\
LLaVA-v1.5-7b$^{\clubsuit}$ & 57.25 & 50.0 & 62.76 & 58.0 & 62.67 & 54.0 & 60.16 & 50.0 & 53.33 & 61.54 & 65.18 & 57.33 \\
LLaVA-v1.5$^{\ddagger\clubsuit}$& 52.64 & 48.0 & 53.10 & 46.67 & 58.67 & 52.67 & 57.72 & 45.33 & 49.33 & 55.38 & 59.82 & 56.67\\ 
\hline
CVLM (LD=0)  & 55.92 & 49.33 & 62.07 & 53.33 & 61.33 & 49.33 & 62.60 & 47.33 & 50.67 & 60.0 & 68.75 & 57.33 \\
CVLM (LD=2) & 57.06 & 53.33 & 62.76 & 56.00 & 66.67 & 52.67 & 59.35 & 46.67 & 49.33 & 63.08 & 63.39 & 66.0\\
CVLM (LD=4) & 56.25 & 50.67 & 61.38 & 53.33 & 60.0 & 52.67 & 58.54 & 47.33 & 50.0 & 63.08 & 64.29 & 64.0\\
CVLM (LD=8) & 59.20 & 55.33 & 62.76 & 54.0 & 61.33 & 56.67 & 65.85 & 52.67 & 56.00 & 66.15 & 66.07 & 61.33\\
CVLM (3M IKPairs, LD=0) & 58.79 & 59.33 & 65.52 & 50.67 & 63.33 & 54.67 & 65.04 & 52.67 & 50.67 & 61.54 & 65.18 & 62.67\\
CVLM (3M IKPairs, LD=2) & 60.33 & 59.33 & 69.66 & 58.0 & 62.67 & 54.67 & 66.67 & 54.67 & 50.0 & 64.62 & 67.86 & 61.33\\
CVLM-624K (LD=0) & \textbf{61.47} & 58.00 & 62.76 & 58.67 & 65.33 & 62.0 & 65.04 & 54.67 & 58.0 & 64.62 & 69.64 & 62.0 \\
CVLM-624K (LD=2) & 60.54 & 58.00 & 62.76 & 58.67 & 64.0 & 59.33 & 67.48 & 53.33 & 56.00 & 61.54 & 70.54 & 58.67 \\

\bottomrule
\end{tabular}
\caption{\textbf{Held-In testing results on OK-VQA with Commonsense Knowledge}. Baseline results are reported by \citet{li2023comprehensive_gpt4v}. Knowledge names are abbreviated due to space limits. and Transportation (VT); Brands, Companies and Products (BCP); Objects, Material and Clothing (OMC); Sports and Recreation (SR); Cooking and Food (CF); Geography, History, Language and Culture (GHLC); People and Everyday Life (PEL); Plants and Animals (PA); Science and Technology (ST); Weather and Climate (WC); and Other. }
\label{tab:okvqa_results}
\end{table*}

\subsection{Baselines}
We mainly compare the proposed method to those current LMMs as follows:
\textbf{BLIP2~\cite{li2023blip2}} is a Large Visual-Language Model that employs a Q-former to integrate visual features into the linguistic space. It provides detailed and accurate descriptions of visual content, effectively bridgiyng the gap between visual perception and linguistic understanding.
\textbf{MiniGPT4~\cite{zhu2023minigpt}} utilizes the same pretrained vision components as BLIP-2, comprising a vision encoder and a Q-Former network. It introduces a single projection layer to align the encoded visual features with the Vicuna~\cite{vicuna2023}.
\textbf{InstructBLIP~\cite{liu2023visual}} utilizes the same Q-former to map visual information into the language space. Its specialization lies in understanding and responding to specific directives related to images, thereby enabling context-aware interactions with visual data.
\textbf{LLaVA~\cite{liu2023improved}} connects the visual encoder with the Vicuna using a simple projection matrix. It could comprehend and generate multimodal content, seamlessly blending text and images for a holistic method of interpreting and generating varied forms of information.

\subsection{Implementation Details}

We utilize the AdamW~\cite{kingma2014adam} optimizer with a cosine learning rate scheduler to train our model. During the pretraining stage of VKA, we first train it on 2 A100 GPUs using a dataset of \textbf{2 million image-knowledge pairs from Wikipedia} with a global batch size of 128 and a base learning rate of 5e-5. Image knowledge processing is shown in the \textit{Appendix}. In the alignment stage, the model is trained on the same 2 million Wikipedia data using 2 A100 GPUs with a global batch size of 32 and a maximum learning rate of 1e-4. For the final stage, we employ LoRA to fine-tune the language model efficiently. In our implementation, we set the rank to 128 and alpha to 256, with a learning rate of 1e-4 for LoRA parameters and the newly added FKA. We use a smaller learning rate of 2e-5 for MLP and VKA.

\subsection{Main Results}

\textbf{Overall Performance}. We present the comparative performance of all models across seven benchmarks tailored for knowledge-based VQA tasks, as detailed in Table \ref{tab:benchmark_result}. Our method exhibits a significant improvement over established baselines, with performance gains of 4\% and 4.5\% on dataset A-OKVQA and TextVQA, respectively. These advancements underscore the effectiveness of our visual-knowledge alignment modules VKA and FKA in bolstering the capabilities of LMMs, particularly evident in the enhancements to LLaVA-v1.5$^\ddagger$. Despite these improvements, our model slightly underperforms in comparison to InstructBLIP on the SEED-Bench, which may be attributed to the larger scale of multimodal instruction tuning data and larger language models (FlanT5-XXL-11B) used by InstructBLIP. 
We also evaluated our model on PoPE and ScienceQA in Table~\ref{tab:pope}. On the ScienceQA dataset, we observe that including VKA results will largely improve the model performance yet FKA may not bring improvement. This may be attributed to the fine-grained features of images in ScienceQA containing less useful information. PoPE is an object recognition benchmark (only need the answer yes or no, like ''Is there an apple in the image'') and the similar performance indicates that introducing VKA and FKA will not affect the basic perception of an image.

\noindent\textbf{Performance on Different Knowledge Categories}. 
We evaluated the performance of our models across 11 detailed categories within the InfoSeek dataset, as outlined in Table \ref{tab:infoseek_results}. This fine-grained analysis reveals that our CVLM significantly outperforms existing models in specific categories, notably Animal, Vehicle, and Sport, showcasing its enhanced understanding and processing capabilities in these knowledge categories.
Moreover, our comprehensive evaluation extends to the OK-VQA testing set, given in Table \ref{tab:okvqa_results}, further highlighting the impact of incorporating visual-knowledge alignment techniques. This strategic integration leads to notable improvements in knowledge-intensive VQA tasks, particularly in SR and PEL domains.
All these results underscore the effectiveness of our approach in leveraging visual knowledge to enrich model performance across a spectrum of knowledge-driven categories. 


\subsection{Ablation Study}
%

\noindent\textbf{Effects of Visual Knowledge Aligner}. To assess the impact of the Visual Knowledge Aligner on model performance, we trained LLaVA-v1.5 using the identical 504K dataset mentioned earlier. As depicted in Table~\ref{tab:benchmark_result}, when compared to the baseline LLaVA-v1.5$^\ddagger$, CVLM(len=0) incorporating the Visual Knowledge Aligner yielded improved results across all benchmarks. Specifically, CVLM(len=0) exhibited the most significant enhancement on SEED-Bench(Spatial), achieving a 4.5\% increase.

\noindent\textbf{Effects of FKA}. Then we study how the FKA influences the model performance. By comparing the experimental results of CVLM(len=0) and CVLM(len=2) on InfoSeek and OK-VQA benchmarks, and CVLM vs. CVLM w/o FKA in Table~\ref{tab:benchmark_result}, we observe that the performance of the proposed method is further improved when FKA is added to the model. The reason for the improvement in our method's performance is that FKA enables the model to perceive fine-grained knowledge information, thereby further enhancing the understanding ability for knowledge-based questions.

\noindent\textbf{Impact of the Size of Visual Knowledge Pairs}. 
To validate the effectiveness of adding more knowledge pre-training data, we increased the amount of Wikipedia knowledge pre-training data from 2M to 3.3M during training VKA with LLM. The experimental results of CVLM (3M IKPairs) are shown in Tables~\ref{tab:benchmark_result}, \ref{tab:infoseek_results} and \ref{tab:okvqa_results}.
We observed that the inclusion of more pretraining knowledge data significantly enhances the model's ability to comprehend knowledge, consequently resulting in higher performance on various tasks.

\noindent\textbf{Analysis of Distillation Vector Length}. Our examination of the optimal distillation vector length, as shown in Tables \ref{tab:infoseek_results} and \ref{tab:okvqa_results}, indicates that increasing distillation vector length does not significantly improve model performance, rather results in performance fluctuations. This suggests that expanding distillation vectors could disturb the structural integrity of large language models, potentially reducing our model's effectiveness, especially in knowledge-dependent tasks.

\noindent \textbf{Impact of introducing more instruction datasets}. As the bottom results shown in Tables~\ref{tab:infoseek_results} and \ref{tab:okvqa_results}, we can see that more instruction data (624k from LLaVA-v1.5) will bring improvement on two knowledge-based VQA datasets, yet it will degrade the performance on some knowledge categories such as Plant. It indicates that introducing more data may not necessarily bring about an overall improvement in performance. Our cognitive mapper method leads to greater performance improvements with a small amount of instruction data, compared to adding more instruction data.

\noindent \textbf{Analysis of the additional computation costs from SAM}. We experimented with 1, 3, 5, and 8 image objects to determine the optimal count that maintains model performance while minimizing computational demand, as shown in Table~\ref{tab:benchmark_result}. Our findings suggest that extracting 3 or 5 regions offers a balanced approach. Increasing the number of image regions tends to introduce visual noise, which can either slightly degrade performance or offer negligible improvements. Furthermore, most evaluation samples require limited fine-grained visual information, featuring fewer objects.

\begin{figure}[t]
    \centering    \includegraphics[width=0.49\textwidth]{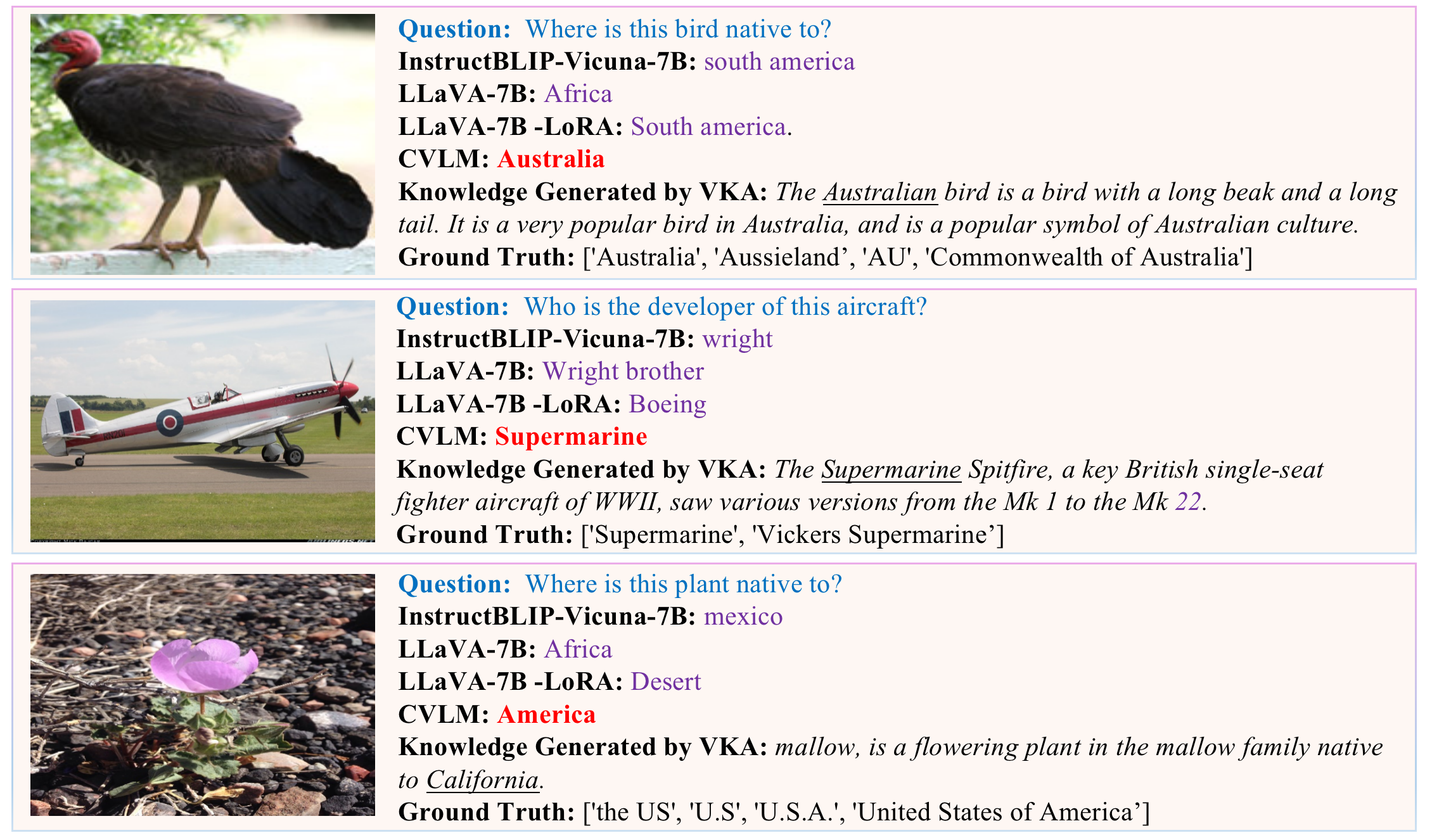}
    \caption{Three cases illustrate the comparative performances of CVLM and other models. Red words represent the correct answer and the purple words show the inaccurate response.}
    \label{fig:case_study}
\end{figure}

\subsection{Case Study}
We present three cases in Figure~\ref{fig:case_study} to thoroughly examine the performance of the models. 
Previous LMMs have been struggling with precise object identification in images and often providing generalized answers to knowledge-based questions. For example, as the case shown in the middle part of Figure~\ref{fig:case_study}, model InstructBLIP-vicuna-7B and LLaVA-7B incorrectly responded with "Wright Brothers" instead of naming the specific aircraft manufacturers shown. Case 3 demonstrated a similar limitation, with LLaVA-7B-LoRA merely predicting "Desert." 
However, employing the VKA and FKA mechanisms, CVLM displays superior capability in discerning crucial elements within images, thus furnishing more precise and contextually relevant responses based on pertinent knowledge.

\section{Conclusion}
In this work, we introduce the Cognitive Visual-Language Mapper (CVLM), an innovative approach that goes beyond the conventional alignment of visual and textual descriptions by incorporating visual-knowledge alignment. Specifically, we have developed a Visual Knowledge Aligner (VKA) that facilitates the projection of visual knowledge by acting as a bridge between the vision encoder and LLM. Additionally, we have integrated a Fine-grained Visual Knowledge Adapter (FKA) during the multimodal instruction tuning stage, which is designed to distil more precise knowledge pertinent to images and instructions. Our experimental findings demonstrate that CVLM outperforms several prominent LMMs that lack visual knowledge alignment. Our ablation studies highlight the effectiveness of VKA and FKA.

\section{Acknowledge}

Thanks for the efforts from reviewers and action editors. This work is supported by grants: Natural Science Foundation of China (No. 62376067).

\section*{Limitations}
Our work has several limitations: 1) The knowledge representations gained by VKA may be inaccurate due to the loss of visual information and errors of knowledge association. Although we introduce large-scale visual knowledge data and the FKA to enhance visual knowledge acquisition, there is still potential to improve the accuracy of visual knowledge alignment.
2) From the experimental results, we observed that the distillation vector length impacts the stability of language models infused with visual knowledge information. Hence, we still need to explore an effective and stable
visual knowledge-enhanced version of CVLM, especially for its FKA component. 3) The generated content may contain some factual errors or toxic statements due to the limitations of LLMs' generation capabilities.
We also hope this work could spark further research on improving visual knowledge alignment during the construction of LMMs.


\bibliography{anthology,custom}
\bibliographystyle{acl_natbib}

\appendix
\section{How Wikipedia knowledge pretraining data Constructed?}

The construction of our Wikipedia knowledge pretraining data leverages images and their corresponding descriptions from approximately 2 million Wikipedia pages. Each page includes multiple images, enriched with Contextual Image Captioning and section summarization to serve as associated knowledge descriptions for the images. This comprehensive dataset is derived from work[1], providing a foundational resource for our pretraining data.
Data Construction Processes are:

\textbf{Data Processing}: The initial step involves reading and parsing the TFRecord file. To manage the vast amount of data efficiently, concurrently, a thread pool is employed to download images based on the parsed image URLs.

\textbf{Filtering}: It was observed that approximately 1\% of the image URLs were invalid. To ensure the quality and integrity of the dataset, these invalid entries were identified and subsequently removed from the dataset.

[1] WIT: Wikipedia-Based Image Text Dataset for Multimodal Multilingual Machine Learning, 2021. https://github.com/google-research-datasets/wit/blob/main/wikiweb2m.md


\end{document}